\documentclass[10pt,twocolumn,letterpaper]{article}

\usepackage{iccv}
\usepackage{times}
\usepackage{epsfig}
\usepackage{graphicx}
\usepackage{amsmath}
\usepackage{amssymb}

\usepackage{enumitem} % for linesep
\usepackage{float} % for subfloats
\usepackage{subfig} % for subfloats
\usepackage[dvipsnames]{xcolor} % for ForestGreen
\usepackage[accsupp]{axessibility}  % Improves PDF readability for those with disabilities.

\usepackage{bbm,bm}
\usepackage{mathtools}

\usepackage[pagebackref=true,breaklinks=true,letterpaper=true,colorlinks,bookmarks=false]{hyperref}

\DeclarePairedDelimiter\norm{\lVert}{\rVert}

\newcommand{\quotes}[1]{``#1"}

\renewcommand{\vec}[1]{\bm{#1}}

\newcommand{\oneNorm}[2][1]{%
    \ensuremath{{}_{\phantom{#1}}\norm{\vec{#2}}_{#1}}%
}
\newcommand{\twoNorm}[2][2]{%
    \ensuremath{{}_{\phantom{#1}}\norm{\vec{#2}}_{#1}}%
}
\newcommand{\new}[1]{\textcolor{black}{#1}}

\DeclareMathOperator*{\argmax}{arg\,max}

\iccvfinalcopy % *** Uncomment this line for the final submission

\def\iccvPaperID{10137} % *** Enter the ICCV Paper ID here

\ificcvfinal\fi

\begin{document}

%%%%%%%%% TITLE
%\title{Quantized Neural Networks for Low-Precision Accumulation\\with Guaranteed Overflow Avoidance}
\title{A2Q: Accumulator-Aware Quantization with Guaranteed Overflow Avoidance}

\author{Ian Colbert, Alessandro Pappalardo, Jakoba Petri-Koenig\\
Advanced Micro Devices, Inc.\\
{\tt\small \{icolbert, alessand, jakobap\}@amd.com}
}

\maketitle
% Remove page # from the first page of camera-ready.
\ificcvfinal\thispagestyle{empty}\fi

%%%%%%%%% ABSTRACT
\begin{abstract}
We present accumulator-aware quantization (A2Q), a novel \new{weight} quantization method designed to train quantized neural networks (QNNs) to avoid overflow when using low-precision accumulators during inference.
A2Q introduces a unique formulation inspired by weight normalization that constrains the $\ell_1$-norm of model weights according to accumulator bit width bounds that we derive.
Thus, in training QNNs for low-precision accumulation, A2Q also inherently promotes unstructured weight sparsity to guarantee overflow avoidance.
\new{We apply our method to deep learning-based computer vision tasks to show that A2Q can train QNNs for low-precision accumulators while maintaining model accuracy competitive with a floating-point baseline.}
{In our evaluations, we consider the impact of A2Q on both general-purpose platforms and programmable hardware.
However, we primarily target model deployment on FPGAs because they can be programmed to fully exploit custom accumulator bit widths.
Our experimentation shows accumulator bit width significantly impacts the resource efficiency of FPGA-based accelerators.
On average across our benchmarks, A2Q offers up to a 2.3x reduction in resource utilization over 32-bit accumulator counterparts with 99.2\% of the floating-point model accuracy.}
%In doing so, we also show that reducing the accumulator bit width translates to increased design efficiency for FPGA-based accelerators.
%On average across our benchmarks, A2Q offers up to a 2.3x reduction in resource utilization over 32-bit accumulator counterparts with 99.2\% of the floating-point model accuracy.
\end{abstract}

%%%%%%%%% BODY TEXT
%\vspace{-0.3cm}
\section{Introduction}
\label{sec:introduction}

Quantization is the process of reducing the range and precision of the numerical representation of data.
%Among the many techniques used to reduce the inference costs of neural networks, integer quantization is one of the most widely applied in practice, usually in exchange for minor reductions in model accuracy~\cite{gholami2021survey, hawks2021ps, hubara2017quantized, zhang2022learning, zhang2021training}.
When applied to the weights and activations of neural networks, integer quantization reduces compute and memory requirements, usually in exchange for minor reductions in model accuracy~\cite{gholami2021survey, hubara2017quantized, jacob2018quantization, zhang2022learning}.
During inference, most of the compute workload is concentrated in operators such as convolutions and matrix multiplications, whose products are typically accumulated into 32-bit registers that we refer to as accumulators.
\new{It has been shown that reducing accumulators to 16 bits on CPUs and ASICs can increase inference throughput and bandwidth efficiency by up to 2x~\cite{de2020quantization, xie2021overflow}, and reducing to 8 bits can improve energy efficiency by over 4x~\cite{ni2021wrapnet}.}
{However, exploiting such an optimization is highly non-trivial as doing so incurs a high risk of overflow.
Due to wraparound two's complement arithmetic, this can introduce numerical errors that degrade model accuracy~\cite{ni2021wrapnet}.}

Previous works have sought to either reduce the risk of overflow~\cite{li2022downscaling, sakr2019accumulation, xie2021overflow} or mitigate its impact on model accuracy~\cite{ni2021wrapnet}.
%However, as discussed in Section~\ref{sec:background}, such approaches only approximate overflow and therefore cannot accurately estimate its impact.
However, such approaches struggle to maintain accuracy when overflow occurs too frequently~\cite{ni2021wrapnet}, and are unable to support applications that {require guaranteed arithmetic correctness}, such as finite-precision fully homomorphic encryption computations~\cite{lou2019she, stoian2023deep}.
Thus, we are motivated to avoid overflow altogether.
As the first principled approach to guarantee overflow avoidance, we provide theoretical motivation in Section~\ref{sec:bounds}, where we derive comprehensive accumulator bit width bounds with finer granularity than existing literature.
In Section~\ref{sec:method}, we present accumulator-aware quantization (A2Q); a novel method designed to train quantized neural networks (QNNs) to use low-precision accumulators during inference without any risk of overflow.
%In Section~\ref{sec:experiments}, we show that our method not only prepares QNNs for low-precision accumulation, but also inherently increases the sparsity of the resulting weights by constraining their $\ell_1$-norm.
In Section~\ref{sec:experiments}, we show that our method not only prepares QNNs for low-precision accumulation, but also inherently increases the sparsity of the weights. % by constraining their $\ell_1$-norm.

%On average across all of our benchmarks, we observe that constraining the hidden layers of QNNs to use 16-bit accumulators without overflow yields up to 91.5\% sparsity with an estimated compression rate of 17.3x while maintaining 99.3\% of the floating-point accuracy.

%To exploit the wider design space exposed by considering low-precision weights, activations, and accumulators, we target model deployment on custom FPGA accelerators rather than general-purpose platforms such as CPUs or GPUs.
While our results have implications for general-purpose platforms such as CPUs and GPUs, we primarily target model deployment on custom FPGA-based inference accelerators.
FPGAs allow bit-level control over every part of a low-precision inference accelerator and can therefore take advantage of custom data types to a greater extent than general-purpose platforms, which are often restricted to power-of-2 bit widths.
In doing so, we show in Section~\ref{sec:experiments} that reducing the bit width of the accumulator can in turn improve the overall trade-off between resource utilization and model accuracy for custom low-precision accelerators.
%On average across all of our benchmarks, we observe that reducing the bit width of the accumulator below 32-bits can reduce resource utilization by up to 2.5x while maintaining 99.3\% of the floating-point accuracy.

To the best of our knowledge, we are the first to explore the use of low-precision accumulators to improve the design efficiency of \new{FPGA-based} QNN inference accelerators.
As such, we integrate A2Q into the open-source Brevitas quantization library~\cite{brevitas} and FINN compiler~\cite{xilinx2023finn} to demonstrate an end-to-end flow for training QNNs for low-precision accumulation and generating custom streaming architectures targeted for AMD-Xilinx FPGAs.

\section{Background}
\label{sec:background}

%While many researchers and practitioners have taken to leveraging reduced precision representations for weights and activations~\cite{gholami2021survey, hubara2017quantized, jacob2018quantization, nagel2021white}, few works have focused attention on reduced precision accumulators~\cite{de2020quantization, sakr2019accumulation, xie2021overflow, ni2021wrapnet}.
%Our work explores the use of weight normalization as a means of constraining weights during QAT for the purpose of avoiding overflow when using low-precision accumulators.
%Here, we provide background related to this objective.

\subsection{Quantization-Aware Training (QAT)}
\label{sec:background_qat}

The standard operators used to emulate quantization during training rely on uniform affine mappings from a high-precision real number to a low-precision quantized number~\cite{jacob2018quantization}.
The quantizer (Eq.~\ref{eq:quantizer}) and dequantizer (Eq.~\ref{eq:dequantizer}) are parameterized by zero-point $z$ and scaling factor $s$.
Here, $z$ is an integer value that ensures that zero is exactly represented in the quantized domain, and $s$ is a strictly positive real scalar that corresponds to the resolution of the quantization function.
Scaled values are rounded to the nearest integers using half-way rounding, denoted by $\lfloor \cdot \rceil$, and elements that exceed the largest supported values in the quantized domain are clipped such that $\textrm{clip}(x; n,p) = \min( \max(x ;n) ;p)$, where $n$ and $p$ depend on the data type of $x$.
For signed integers of bit width $b$, we assume $n=-2^{b-1}$ and $p=2^{b-1} - 1$.
For unsigned integers, we assume $n=0$ and $p=2^b - 1$ when unsigned.
\begin{align}
\textrm{quantize}(x;s,z) & := \textrm{clip}(\left\lfloor \frac{x}{s} \right\rceil + z; n, p) \label{eq:quantizer} \\
\textrm{dequantize}(x;s,z) & := s \cdot (x - z) \label{eq:dequantizer}
\end{align}

It has become common to use unique scaling factors for each of the output channels of the learned weights to adjust for varied dynamic ranges~\cite{nagel2019data}.
However, extending this strategy to activations requires either storing partial sums or introducing additional control logic.
As such, it is standard practice to use per-tensor scaling factors for activations and per-channel scaling factors on the weights.
It is also common to constrain the weight quantization scheme such that $z=0$~\cite{gholami2021survey}.
Eliminating these zero points reduces the computational overhead of cross-terms during integer-only inference~\cite{jain2020trained}.
During training, the straight-through estimator (STE)~\cite{bengio2013estimating} is used to allow local gradients to permeate the rounding function such that $\nabla_x \lfloor x \rceil = 1$ everywhere, where $\nabla_x$ denotes the local gradient with respect to $x$.

\begin{figure}[t!]
\centering
\includegraphics[width=0.75\linewidth]{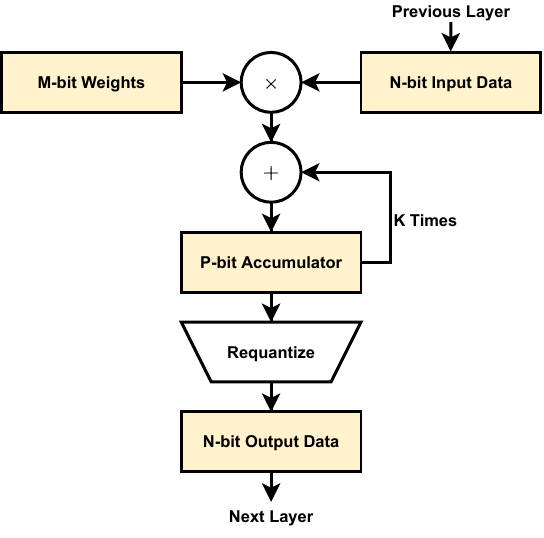}
\caption{A simplified illustration of fixed-point arithmetic in neural network inference.
%Quantized weights are frozen during inference.
%Input/output data is dynamic and thus, scaled then clipped as activations are passed through the network.
The accumulator bit width ($P$) needs to be wide enough to fit the dot product between the $M$-bit weight vector and the $N$-bit input vector, which are assumed to both be $K$-dimensional.}
\label{fig:accumulator_flow_diagram}
\end{figure}

\begin{figure}[h]
\includegraphics[width=\linewidth]{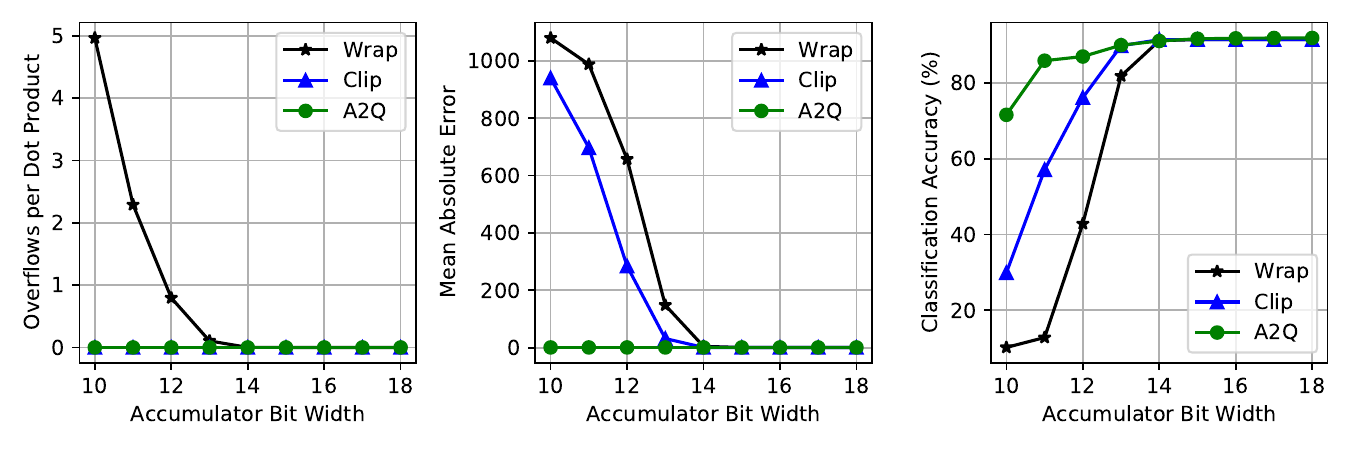}
\caption{We evaluate the impact of overflow as we reduce the accumulator bit width using a 1-layer QNN trained to classify binary MNIST~\cite{deng2012mnist} images using 8-bit weights.
We show that using A2Q (\textcolor{ForestGreen}{\textbf{green dots}}) to avoid overflow significantly improves model accuracy over both wraparound arithmetic (\textbf{black stars}) and clipping (\textcolor{blue}{\textbf{blue triangles}}) when using extremely low-precision accumulators.}
\label{fig:motivation}
\end{figure}

\subsection{Low-Precision Accumulation}
\label{sec:background_lpa}

%\begin{figure*}[t!]
%\subfloat[FINN Framework Overview]{\includegraphics[width=0.6\linewidth]{finn-flow.pdf}}
%\subfloat[MVAU]{\includegraphics[width=0.4\linewidth]{mvau.pdf}}
%\caption{We adapt images from~\cite{finn, blott2018finn} to provide: (a) an overview of the FINN framework; and (b) an abstraction of the matrix-vector-activation unit (MVAU), which is one of the primary building blocks used by the FINN compiler to generate custom streaming architectures.}
%\label{fig:finn}
%\end{figure*}

As activations are propagated through the layers of a QNN, the intermediate partial sums resulting from convolutions and matrix multiplications are typically accumulated in a high-precision register before being requantized and passed to the next layer, as depicted  in Fig.~\ref{fig:accumulator_flow_diagram}.
Reducing the precision of the accumulator incurs a high risk of overflow which, due to wraparound two's complement arithmetic, introduces numerical errors that can degrade model accuracy~\cite{ni2021wrapnet}.
%To demonstrate this, we train a 1-layer QNN to classify binary MNIST~\cite{deng2012mnist} images using 8-bit weights. 
As shown in Fig.~\ref{fig:motivation}, the rate of overflows per dot product often grows exponentially as the accumulator bit width is reduced (please see Appendix~\ref{sec:motivation_exp_details} for details).
The increased overflow rate introduces numerical errors that proportionally increase the mean absolute error on the logits, decreasing classification accuracy.
The industry standard for avoiding overflow is to either use high-precision accumulators or na\"ively saturate values as they are accumulated.
However, such clipping can still: (1) introduce numerical errors that cascade when propagated through a QNN; and (2) require additional logic that can break associativity while adding to latency and area requirements~\cite{xilinx2023saturation}.
In breaking associativity, the final result of the dot product is made dependent on the order of additions. This can introduce non-deterministic errors when modern processors use optimizations that improve hardware utilization by re-ordering operations~\cite{gale2020sparse, kaeli2015heterogeneous} \new{(please see Appendix~\ref{sec:breaking_associativity} for details)}. % (\textit{e.g.}, out-of-order execution~\cite{kaeli2015heterogeneous} or cache swizzling~\cite{gale2020sparse}).
This further motivates us to train QNNs to completely avoid overflow rather than simply reduce its impact on model accuracy.
%In doing so, A2Q circumvents these numerical errors and improves performance when using extremely low-precision accumulators.
{This way, A2Q entirely circumvents these numerical errors while delivering improved resource efficiency and increased accuracy.}

\section{Accumulator Bit Width Bounds}
\label{sec:bounds}

Figure~\ref{fig:accumulator_flow_diagram} illustrates a simplified abstraction of accumulation in QNN inference.
%As activations are propagated through the layers, the intermediate partial sums resulting from operations such as convolutions or matrix multiplications are accumulated into a register before being requantized and passed to the next layer.
To avoid overflow, the register storing the accumulated values needs to be wide enough to not only contain the result of the dot product, but also all intermediate partial sums.

Consider the dot product of input data $\vec{x}$ and learned weights $\vec{w}$, which are each $K$-dimensional vectors of integers.
Let $y$ be the scalar result of their dot product given by Eq.~\ref{eq:dot_product}, where $x_i$ and $w_i$ denote element $i$ of vectors $\bm{x}$ and $\bm{w}$, respectively.
Since the representation range of $y$ is bounded by that of $\vec{x}$ and $\vec{w}$, we use their ranges to derive lower bounds on the bit width $P$ of the accumulation register, or accumulator.
\begin{equation}
y = \textstyle \sum_{i=1}^K x_i w_i
\label{eq:dot_product}
\end{equation}

It is common for input data to be represented with unsigned integers either when following activation functions with non-negative dynamic ranges (\textit{e.g.}, rectified linear units, or ReLUs), or when an appropriate zero point is adopted (\textit{i.e.}, asymmetric quantization).
Otherwise, signed integers are used.
Since weights are most often represented with signed integers, we assume the accumulator is always signed in our work.
Therefore, given that the scalar result of the dot product between $\vec{x}$ and $\vec{w}$ is a $P$-bit integer defined by Eq.~\ref{eq:dot_product}, it follows that $\textstyle \sum_{i=1}^K x_i w_i$ is bounded such that:
\begin{equation}
-2^{P - 1} \leq \textstyle \sum_{i=1}^K x_i w_i \leq 2^{P - 1} - 1
\label{eq:basic_bound_on_dot_prod}
\end{equation}
To satisfy both sides of this double inequality, it follows that $\vert \textstyle \sum_{i=1}^K x_i w_i \vert \leq 2^{P - 1} - 1$.
However, the accumulator needs to be wide enough to not only store the final result of the dot product, but also all intermediate partial sums.

Since input data is not known \textit{a priori}, our bounds must consider the worst-case values for every MAC.
Thus, because the magnitude of the sum of products is upper-bounded by the sum of the product of magnitudes, it follows that if $\textstyle \sum_{i=1}^K \vert x_i \vert \vert w_i \vert \leq 2^{P - 1} - 1$, then the dot product between $\vec{x}$ and $\vec{w}$ fits into a $P$-bit accumulator without numerical overflow, as shown below.
\begin{equation}
\vert \textstyle \sum_i x_i w_i \vert \leq \textstyle \sum_i \vert x_i w_i \vert \leq \textstyle \sum_i \vert x_i \vert \vert w_i \vert \leq 2^{P - 1} - 1
\label{eq:mag_identities}
\end{equation}

\subsection{Deriving Lower Bounds Using Data Types}
\label{sec:datatype_bounds}

The worst-case values for each MAC can na\"ively be inferred from the representation range of the data types used.
When $x_i$ and $w_i$ are signed integers, their magnitudes are bounded such that $\vert x_i \vert \leq 2^{N - 1}$ and $\vert w_i \vert \leq 2^{M - 1}$, respectively.
In scenarios where $x_i$ is an unsigned integer, the magnitude of each input value is upper-bounded such that $\vert x_i \vert \leq 2^N - 1$;
however, we consider the case where $\vert x_i \vert \leq 2^N$ to simplify our derivation\footnote{Note that our simplification of the upper bound for unsigned input data means that the lower bound on the accumulator is not as tight as possible, but it does not compromise overflow avoidance.}.
Combining these upper bounds, it follows that $\vert x_i \vert \leq 2^{N - \mathbbm{1}_\text{signed}(\vec{x})}$, where $\mathbbm{1}_\text{signed}(\bm{x})$ is an indicator function that returns 1 if and only if $\bm{x}$ is a vector of signed integers.

Building from Eq.~\ref{eq:mag_identities}, it follows that the sum of the product of the magnitudes is bounded such that:
\begin{equation}
\textstyle  \sum_{i=1}^K \vert x_i \vert \vert w_i \vert \leq K \cdot 2^{N + M - 1 - \mathbbm{1}_\text{signed}(\bm{x})} \leq 2^{P - 1} - 1
\label{eq:unfinished_datatype_bound}
\end{equation}
Taking the log of both sides of  Eq.~\ref{eq:unfinished_datatype_bound}, we can derive a lower bound on the accumulator bit width $P$:
\begin{equation}
\log_2 \left(2^{\log_2(K) + N + M - 1 - \mathbbm{1}_\text{signed}(\bm{x})} + 1 \right) + 1 \leq P
\end{equation}
This simplifies to the following lower bound on $P$:
\begin{align}
P & \geq \alpha + \phi(\alpha) + 1 \label{eq:datatype_lower_bound}\\
\alpha & = \log_2(K) + N + M - 1 - \mathbbm{1}_\text{signed}(\bm{x}) \label{eq:datatype_alpha}\\
\phi(\alpha) & = \log_2(1 + 2^{-\alpha}) \label{eq:phi_alpha}
\end{align}

\begin{figure}[t!]
\centering
\subfloat[Using Data Type]{\includegraphics[width=0.5\linewidth]{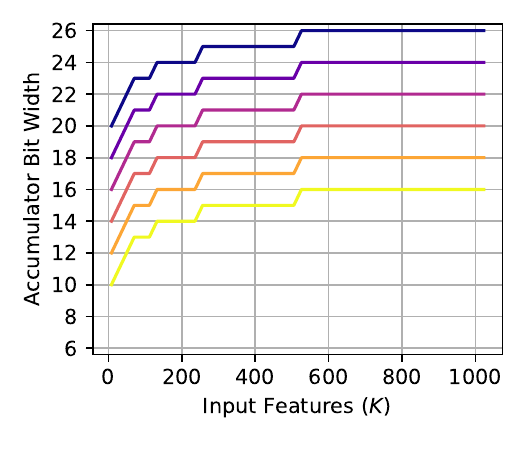}}
\subfloat[Using Weights]{\includegraphics[width=0.5\linewidth]{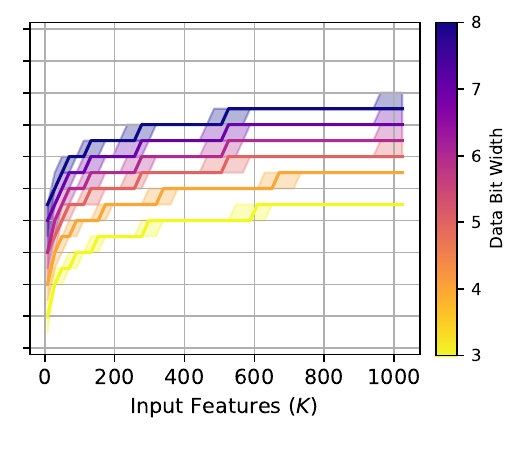}}
\caption{We visualize the differences between our accumulator bit width bounds as we vary the size of the dot product ($K$) as well as the bit width of both the weights ($M$) and inputs ($N$), which we jointly refer to as \quotes{data bit width.}}
\label{fig:acc_bit_width_bounds}
\end{figure}

In Fig.~\ref{fig:acc_bit_width_bounds}a, we visualize this bound assuming that $\vec{x}$ is a vector of unsigned integers.
There, we show how the lower bound on the accumulator bit width increases as we vary the size of the dot product ($K$) as well as the bit width of both the weights and input activations.

\subsection{Deriving Lower Bounds Using Weights}
\label{sec:weight_bounds}

Since learned weights are frozen during inference time, we can use knowledge of their magnitudes to derive a tighter lower bound on the accumulator bit width.
Building again from Eq.~\ref{eq:mag_identities}, the sum of the product of magnitudes is bounded by Eq.~\ref{eq:first_weight_bound_eq}, where $\Vert \bm{w} \Vert_1$ denotes the standard $\ell_1$-norm over vector $\bm{w}$.
\begin{equation}
\textstyle \sum_{i=1}^K \vert x_i \vert \vert w_i \vert \leq 2^{N - \mathbbm{1}_\text{signed}(\bm{x})} \cdot \Vert \bm{w} \Vert_1 \leq 2^{P - 1} - 1
\label{eq:first_weight_bound_eq}
\end{equation}
Accounting for $\oneNorm{w}$ in our derivation allows us to tighten the lower bound on $P$ as follows:
\begin{align}
P & \geq \beta + \phi(\beta) + 1\\
\beta &=  \log_2(\Vert \bm{w} \Vert_1) + N - \mathbbm{1}_\text{signed}(\bm{x}) \label{eq:weight_bound}\\
\phi(\beta) & = \log_2(1 + 2^{-\beta})
\end{align}

In Fig.~\ref{fig:acc_bit_width_bounds}b, we visualize this bound, again assuming that $\vec{x}$ is a vector of unsigned integers.
Because Eq.~\ref{eq:weight_bound} is dependent on the values of the learned weights, we randomly sample each $K$-dimensional vector from a discrete Gaussian distribution and show the median accumulator bit width along with the minimum and maximum observed over 1000 random samples.
Across $K$, $M$, and $N$, we visualize how using knowledge of the weights provides a tighter lower bound on the accumulator bit width than using data types.

\section{A2Q: Accumulator-Aware Quantization}
\label{sec:method}

To train QNNs to use low-precision accumulators without overflow, we use weight normalization as a means of constraining learned weights $\vec{w}$ to satisfy the bound derived in Section~\ref{sec:weight_bounds}.
Building from Eq.~\ref{eq:first_weight_bound_eq}, we transform our lower bound on accumulator bit width $P$ to be the upper bound on the $\ell_1$-norm of $\vec{w}$ given by Eq.~\ref{eq:weight_bound_reparameterized}.
Note that because each output neuron requires its own accumulator, this upper bound needs to be enforced channelwise.
\begin{equation}
\Vert \bm{w} \Vert_1 \leq \left( 2^{P - 1} - 1 \right) \cdot 2^{\mathbbm{1}_\text{signed}(\bm{x}) - N}
\label{eq:weight_bound_reparameterized}
\end{equation}

\subsection{Constructing Our Quantization Operator}

Weight normalization, as originally proposed by Salimans \textit{et al.}~\cite{salimans2016weight}, reparameterizes each weight vector $\bm{w}$ in terms of a parameter vector $\bm{v}$ and a scalar parameter $g$ as given in Eq.~\ref{eq:standard_weight_norm}, where $\norm{\vec{v}}_2$ is the Euclidean norm of the $K$-dimensional vector $\bm{v}$~\cite{salimans2016weight}.
This simple reparameterization fixes the Euclidean norm of weight vector $\bm{w}$ such that $\norm{\bm{w}}_2 = g$, which enables the magnitude and direction to be independently learned.
\begin{equation}
	\bm{w} = g \cdot \frac{\vec{v}}{\twoNorm{v}}
	\label{eq:standard_weight_norm}
\end{equation}

Inspired by this formulation, we reparameterize our quantizer such that each weight vector $\vec{w}$ is represented in terms of parameter vectors $\vec{v}$ and $\vec{g}$.
Similar to the standard weight normalization formulation, this reparameterization decouples the norm from the weight vector; however, unlike the standard formulation, our norm is learned for each output channel rather than per-tensor.
To enforce our constraint during QAT, we also replace the per-tensor $\ell_2$-norm with a per-channel $\ell_1$-norm.
%For a given layer with $C$ output channels, we replace the per-tensor $\ell_2$-norm of the standard formulation (Eq.~\ref{eq:standard_weight_norm}) with a per-channel $\ell_1$-norm.
This reparameterization, given by Eq.~\ref{eq:our_weight_normalization}, allows for the $\ell_1$-norm of weight vector $\vec{w}$ to be independently learned per-channel such that $g_i = \Vert \bm{w}_i \Vert_1$ for all $i \in \{1, \cdots, C \}$.
Here, $\vec{w}_i$ denotes the weights of channel $i$ and $g_i$ denotes element $i$ in parameter vector $\vec{g}$ for a given layer with $C$ output channels.
\begin{equation}
\vec{w}_i = g_i \cdot \frac{\vec{v_i}}{\oneNorm{v_i}}	\quad \forall ~i \in \{1, \cdots, C \}
\label{eq:our_weight_normalization}
\end{equation}

Similar to the standard weight quantizer, our weight normalization-based quantization operator relies on a uniform affine mapping from the high-precision real domain to the low-precision quantized domain using learned per-channel scaling factors $\vec{s}=\{s_i\}^C_{i=1}$.
Thus, by constraining $g_i$ to satisfy Eq.~\ref{eq:constrain_g}, 
we can learn quantized weights that satisfy our accumulator bit width bound and avoid overflow.
\begin{equation}
g_i \leq s_i \cdot \left( 2^{P - 1} - 1 \right) \cdot 2^{\mathbbm{1}_\text{signed}(\bm{x}) - N}
\label{eq:constrain_g}
\end{equation}

Below, we articulate our weight normalization-based quantizer.
For clarity and convenience of notation, we consider a layer with one output channel (\textit{i.e.}, $C=1$) such that parameter vectors $\vec{g}=\{g_i\}^C_{i=1}$ and $\vec{s}=\{s_i\}^C_{i=1}$ can be represented as scalars $g$ and $s$, respectively.
\begin{equation}
\textrm{quantize}(\bm{w};s,z) := \textrm{clip}(\left\lfloor \frac{g}{s} \frac{\vec{v}}{\oneNorm{v}} \right\rfloor + z; n, p) \label{eq:quantizer_ours}
\end{equation}

We construct \textbf{accumulator-aware quantization (A2Q)} from our weight normalization-based quantizer (Eq.~\ref{eq:quantizer_ours}) and the standard dequantizer (Eq.~\ref{eq:dequantizer}).
%The resulting quantization operator for A2Q is given by Eq.~\ref{eq:qat_weight_norm}, where $n$ and $p$ depend on the representation range of weight bit width $M$.
During training, A2Q applies the following four elementwise operations in order: scale, round, clip, then dequantize.
As is standard practice, we eliminate the zero points in our mapping such that $z=0$.
We use an exponential parameterization of both the scaling factor $s = 2^{d}$ and the norm parameter $g = 2^t$, where $d$ and $t$ are both log-scale parameters to be learned through stochastic gradient descent.
This is similar to the work of~\cite{jain2020trained} with the difference that we consider the more common scenario of floating-point scaling factors.
%This is similar to the work of~\cite{jain2020trained} with the caveat that we remove integer power-of-2 constraints, which provide no added benefit to the streaming architectures generated by FINN as floating-point scaling factors can be absorbed into the threshold logic via mathematical manipulation~\cite{blott2018finn}.
The scaled tensors are then rounded towards zero, which we denote by $\lfloor \cdot \rfloor$, to prevent any upward rounding that may cause the norm to increase past our constraint\footnote{It is important to note that rounding towards zero is functionally different from floor or ceiling rounding~\cite{loroch2017tensorquant}.}.
Note that this is another difference from the conventional quantization operators, which primarily use half-way rounding~\cite{gholami2021survey, jain2020trained}.
Finally, once scaled and rounded, the elements in the tensor are then clipped and dequantized.
To update learnable parameters throughout training, we use STE~\cite{bengio2013estimating} as is common practice.
\begin{align}
q(\bm{w} ; s) & := \text{clip}\left( \left\lfloor \frac{g}{s} \frac{\vec{v}}{\oneNorm{v}} \right\rfloor ; n, p \right) \cdot s \label{eq:qat_weight_norm} \\
\text{ where } & s = 2^{d} \\
\text{ and } & g = 2^{\min(T, t)} \label{eq:min_g_t_T} \\
\text{ and } & T = \mathbbm{1}_\text{signed}(\bm{x}) + \log_2(2^{P - 1} - 1) + d - N \label{eq:T}
\end{align}

We apply A2Q to only the weights of a QNN.
To avoid $t$ getting stuck when $t > T$, we introduce the following regularization penalty for the $l$-th layer of the network: $R_l = \textstyle \sum_i \max \{ t_i - T_i, 0 \}$.
This penalty is imposed on every hidden layer and combined into one regularizer: $\mathcal{L}_\text{reg} = \sum_l R_l$.
When quantizing our activations, we use the standard quantization methods discussed in Section~\ref{sec:background_qat}.
All activations that follow non-negative functions (\textit{i.e.}, ReLU) are represented using unsigned integers, otherwise they are signed.

\section{Experiments}
\label{sec:experiments}

\new{Without guaranteed overflow avoidance, one cannot reliably design hardware accelerators around low-precision accumulators.
Therefore, in our experiments, we do not compare against methods that cannot provide such guarantees.
Given that, we consider two scenarios in the following evaluations.}
In Section~\ref{sec:scenario_1}, we optimize QNNs for accumulator-constrained processors, where the goal is to maximize task performance given a user-defined accumulator bit width.
Such a scenario is a direct application of our method and has implications for both accelerating inference on general-purpose platforms~\cite{de2020quantization, ni2021wrapnet, xie2021overflow} and reducing the computational overhead of homomorphic encryption arithmetic~\cite{lou2019she, stoian2023deep}.
In Section~\ref{sec:scenario_2}, we optimize QNNs for overall resource utilization within a hardware-software (HW-SW) co-design setting.
In this scenario, the goal is to maximize task performance given a user-defined hardware resource budget.
Our experiments show that including the accumulator bit width as part of the design space can improve the trade-off between resources and accuracy.
We target model deployment on custom FPGA-based accelerators, rather than CPUs or GPUs, as they allow bit-level control over every part of the network and can therefore take advantage of custom data types to a greater extent.
To do so, we adopt FINN~\cite{blott2018finn, finn}, an open-source framework designed to generate specialized streaming architectures for QNN inference acceleration on AMD-Xilinx FPGAs.
We build on top of FINN v0.8.1~\cite{xilinx2023finn} and open-source our A2Q implementation\footnote{\url{https://github.com/Xilinx/brevitas/tree/master/src/brevitas_examples}} as part of Brevitas v0.10~\cite{brevitas}.

\subsection{Experiment Setup}
\label{sec:exp_setup}

We apply A2Q to the following two computer vision tasks: (1) image classification on CIFAR10~\cite{krizhevsky2009learning} using MobileNetV1~\cite{howard2017mobilenets} and ResNet18~\cite{he2016deep}; and (2) single-image super resolution on BSD300~\cite{MartinFTM01} using ESPCN~\cite{shi2016real} and UNet~\cite{ronneberger2015u}.
For each model, we measure task performance over the test dataset, where image classification and single-image super resolution models are evaluated using top-1 classification accuracy and peak signal-to-noise ratio (PSNR), respectively.
We include more details on model and training settings in Appendix~\ref{sec:exp_details}.

Throughout our experiments, we constrain our quantization design space to uniform-precision models.
For every hidden layer in each network, we enforce the same weight, activation, and accumulator bit width, respectively denoted as $M$, $N$, and $P$.
We perform a grid search over our quantization design space, focusing our attention on weight and activation bit widths between 5 and 8 bits.
Doing so allows even comparisons across bit widths, as reducing the precision below 5 bits often requires unique hyperparameters to maximize performance.
For each weight and activation bit width combination, we calculate the largest lower bound on the accumulator bit width for each model.
In a model with $L$ layers, this is determined by the data type bound of the layer with the largest dot product size $K^*$, where $K^* = \argmax_{K_l} \{ K_l \}_{l=0}^L $.
Using this to guide our grid search over $P$ for each model, we evaluate up to a 10-bit reduction in the accumulator bit width to create a total of 160 configurations per model.

%Although automated mixed-precision quantization algorithms have been explored~\cite{dong2019hawqm wang2019haq}...

\subsection{Optimizing for Accumulator Constraints}
\label{sec:scenario_1}

\begin{figure*}[t!]
\centering
\includegraphics[width=\linewidth]{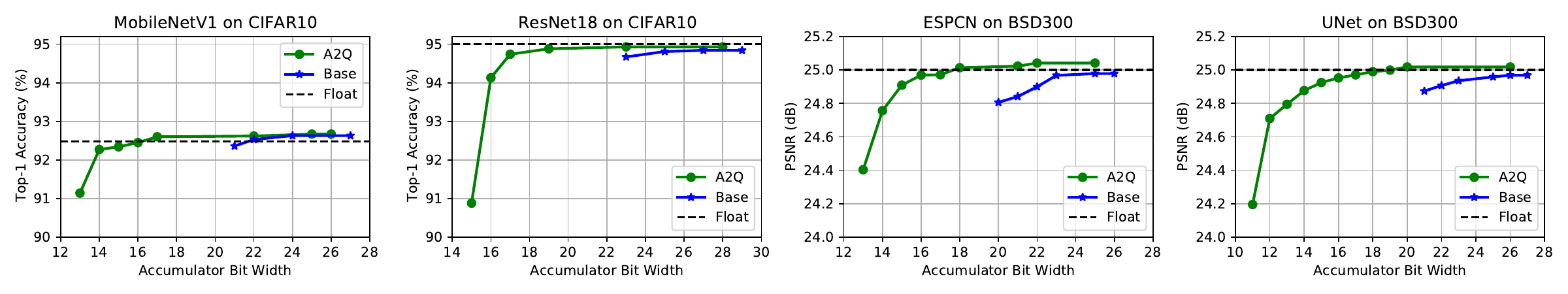}
\caption{
We visualize the trade-off between accumulator bit width and task performance using Pareto frontiers.
%We use $P^*$ to denote the largest accumulator bit width allowed across all layers in the network.
We observe that A2Q (\textcolor{ForestGreen}{\textbf{green dots}}) dominates the baseline QAT (\textcolor{blue}{\textbf{blue stars}}) in all benchmarks, showing that we can reduce the accumulator bit width without sacrificing significant model performance even with respect to a floating-point baseline.
}
\label{fig:pareto_fronts_adt}
\end{figure*}

To the best of our knowledge, A2Q is the first to allow a user to train a QNN to avoid overflow by only specifying a target accumulator bit width $P$.
As an alternative, a designer can choose to heuristically reduce data bit widths based on our data type bound given by Eq.~\ref{eq:datatype_lower_bound}. 
Such an approach would still guarantee overflow avoidance for a user-defined $P$, but is a limited and indirect means of enforcing such a constraint.
By heuristically manipulating data bit widths, the minimum attainable $P$ is bounded by both the quantization design space and the architecture of the QNN because it is a function of $M$, $N$, and the size of the dot product $K$.
Conversely, A2Q exposes $P$ as an independent variable to be directly specified in the design, orthogonal to the rest of the design space.
To compare the performance of models trained with A2Q against the baseline heuristic approach, we vary $M$, $N$, and $P$ across our benchmark models.
We visualize this comparison as a Pareto frontier in Fig.~\ref{fig:pareto_fronts_adt} and provide the floating-point task performance as reference.
For each model and each algorithm, the Pareto frontier shows the maximum observed task performance for a given target accumulator bit width $P$.

It is important to note that, while this is not a direct comparison against the algorithm proposed by~\cite{de2020quantization}, the experiment is similar in principle.
Unlike~\cite{de2020quantization}, we use the more advanced quantization techniques detailed in Section~\ref{sec:background_qat} and constrain our quantization design space to uniform-precision models.
Within this design space, we observe that A2Q can push the accumulator bit width lower than what is attainable using current methods while also maintaining task performance.
Furthermore, most models show less than a 1\% performance drop with a 16-bit accumulator, which is  often the target bit width for low-precision accumulation in general-purpose processors~\cite{de2020quantization, li2022downscaling, xie2021overflow}.

\subsubsection{Accumulator Impact on Model Sparsity}
\label{sec:sparsity}

\begin{figure}[t!]
\centering
\includegraphics[width=\linewidth]{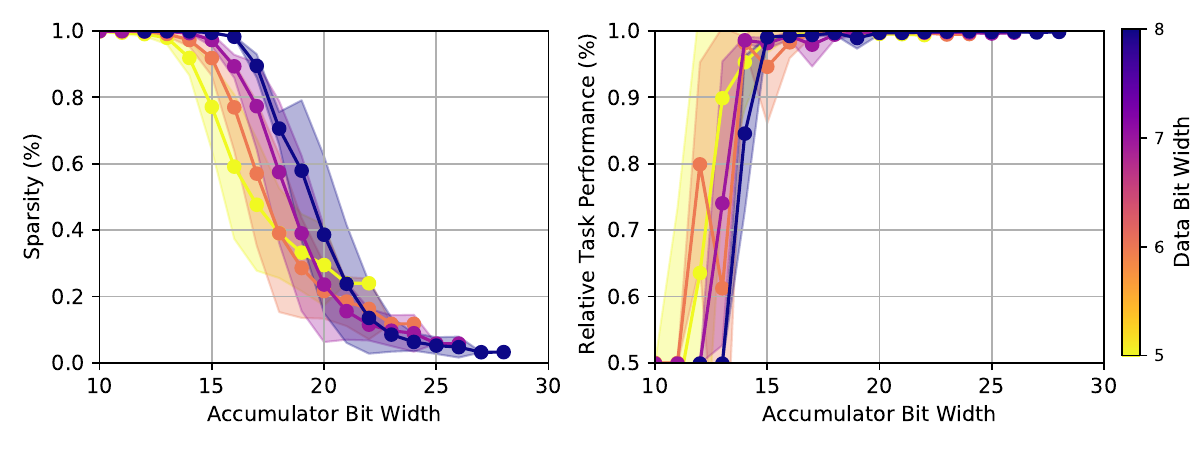}
\caption{As a result of our $\ell_1$-norm constraints, reducing the accumulator bit width exposes opportunities to exploit unstructured sparsity (left) without sacrificing model accuracy relative to the floating-point baseline (right).
{We observed that these trends were similar for each model; thus, to simplify our analysis, we provide the average and standard deviation calculated over all of our benchmark models.}}
\label{fig:sparsity}
\end{figure}

Given a target accumulator bit width $P$, our quantization method constrains the $\ell_1$-norm of model weights according to the upper bound given by Eq.~\ref{eq:weight_bound_reparameterized}.
Consequently, these constraints exponentially tighten as $P$ is reduced (see Eqs.~\ref{eq:constrain_g} and~\ref{eq:T}).
Previous work has studied the use of $\ell_1$-norm weight regularization as a means of introducing sparsity~\cite{chao2020directional, yang2019structured}.
We observe A2Q to inherently have a similar effect when training QNNs for low-precision accumulation.

In Fig.~\ref{fig:sparsity}, we visualize how the sparsity and relative task performance are affected by reductions to $P$.
We use the models from our grid search described in Section~\ref{sec:exp_setup}, but focus on configurations where the weight and input activation bit widths are the same (\textit{i.e.}, $M = N$) to simplify our analysis.
For each accumulator bit width, we plot the average sparsity and relative task performance observed across all of our benchmark models and provide the standard deviation.
On average, we observe that constraining the hidden layers of our QNNs to use less than 32-bit accumulators yields up to 92\% unstructured weight sparsity while maintaining 99.2\% of the floating-point model accuracy.

\subsection{Optimizing for Resource Utilization}
\label{sec:scenario_2}

\begin{figure*}[t!]
\centering
\includegraphics[width=\linewidth]{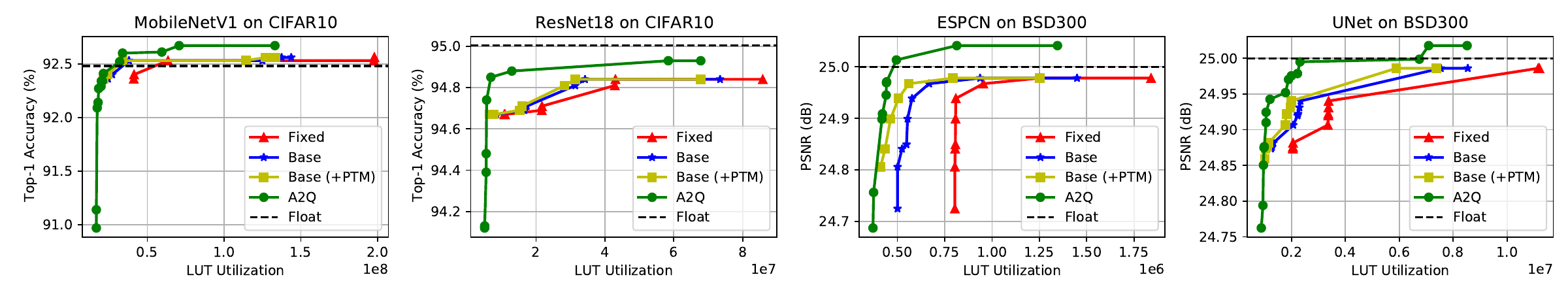}
\caption{We visualize the trade-off between resource utilization and model accuracy using Pareto frontiers.
We compare A2Q (\textcolor{ForestGreen}{\textbf{green dots}}) to baseline QAT with: (1) a fixed accumulator bit width $P$ (\textcolor{red}{\textbf{red triangles}}); (2) layer-wise selection of $P$ using data type bounds (\textcolor{blue}{\textbf{blue dots}}); and (3) post-training minimization (PTM) of $P$ using weight values (\textcolor{Dandelion}{\textbf{yellow squares}}).
We observe that A2Q provides a better trade-off between LUT utilization and task performance than existing baselines.}
\label{fig:pareto_fronts_lut}
\end{figure*}

To evaluate the trade-offs between resource utilization and task performance within our quantization design space, we target QNN model deployment on AMD-Xilinx FPGAs.
We use the FINN framework to generate specialized hardware architectures that are individually customized for the network topology and data types used.
This exposes control over the accumulators used in each layer so that we can take advantage of custom data types in our experiments.
At the core of FINN is its compiler, which typically relies on FPGA look-up tables (LUTs) to perform multiply-and-accumulates (MACs) at low precision.
In such scenarios, LUTs are often the resource bottleneck for the low-precision inference accelerators it generates.
Therefore, we simplify our analysis by configuring the FINN compiler to assume that LUTs are the only type of resources available, and we leverage the LUT utilization estimates for each of QNN trained in our grid search.
We include additional details on FINN in Appendix~\ref{sec:finn}.

Previous work has shown that reducing the precision of weights and activations provides resource utilization savings in exchange for minor reductions to model accuracy~\cite{finn}.
We observe that adding the accumulator bit width to the design space can improve this trade-off.
To demonstrate the impact, we consider four HW-SW co-design settings.
First, we consider a fixed accumulator bit width and remove $P$ from the quantization design space discussed in Section~\ref{sec:exp_setup}.
We use the baseline QAT to train each model and configure the generated accelerator to use a constant 32-bit accumulator for all layers.
Second, we again use the baseline QAT, but configure FINN to use the minimum accumulator bit width $P$ according to the data type bound (Eq.~\ref{eq:datatype_lower_bound}) per-layer.
Third, we again use the baseline QAT, but configure the FINN compiler to further minimize $P$ for each layer according to the $\ell_1$-norm of the final weight values post-training (Eq.~\ref{eq:weight_bound}).
Finally, we evaluate the end-to-end design flow when using A2Q to train QNNs for a specified weight, activation, and accumulator bit width.
For each co-design setting, we visualize the trade-offs between resource utilization and task performance as a Pareto frontier in Fig.~\ref{fig:pareto_fronts_lut} and provide the floating-point task performance for reference.
For each model and each co-design setting, the Pareto frontier shows the maximum observed task performance for a given total LUT utilization.

As expected, we observe that the layer-wise minimization of the accumulator bit width provides a better resource-to-accuracy trade-off than using a fixed 32-bit accumulator.
We also observe that post-training minimization of $P$ according to the final weight values provides consistent improvements over the data type bounds.
Furthermore, our results show that using A2Q to train models for reduced accumulator bit widths provides a dominant Pareto frontier across all models.
For the tasks at hand, A2Q pushes the best attainable task performance above the baselines.
Thus, for a given target accuracy or resource budget, A2Q can offer a better trade-off between LUT utilization and task performance, confirming the benefits of including the accumulator bit width in the overall HW-SW co-design space.

\subsubsection{Evaluating Resource Savings}
\label{sec:resource_savings}

\begin{figure*}[t!]
\centering
\includegraphics[width=\linewidth]{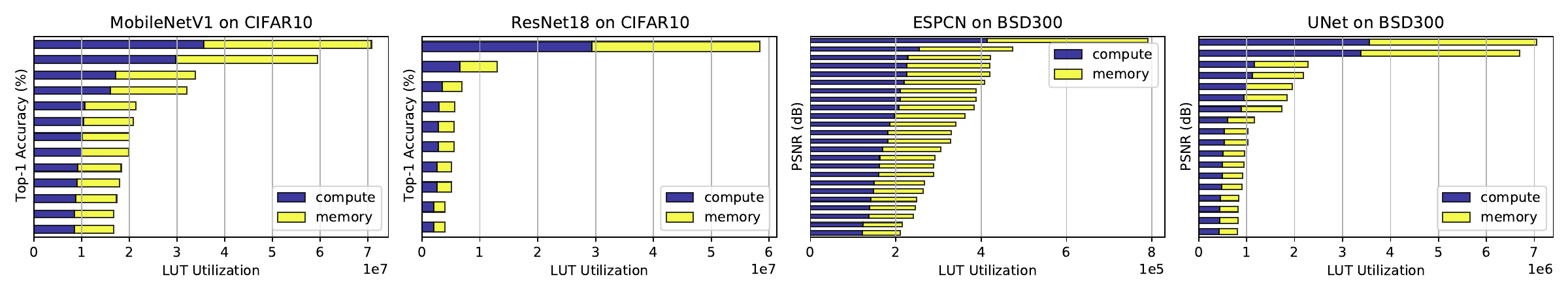}
\caption{We break down LUT utilization for each of our Pareto optimal models from Fig.~\ref{fig:pareto_fronts_lut}.}
\label{fig:pareto_fronts_breakdown}
\end{figure*}

Because we configure the FINN compiler to use LUTs wherever possible, we provide a deeper analysis of where the resource savings come from.
To do so, we separate LUT utilization into compute and memory resources but ignore control flow overhead, which remains constant for each network because neural architecture is not impacted by the data types used.
%For compute, we aggregate the LUTs used for adder trees, MACs, and comparison logic.
%For memory, we aggregate the LUTs used to store weights, thresholds\footnote{Thresholds are used to compute quantized monotonic activation functions, as further described in Appendix~\ref{sec:finn}}, and intermediate activations.
In Fig.~\ref{fig:pareto_fronts_breakdown}, we visualize this break down for each of the Pareto optimal models that correspond to the A2Q Pareto frontier provided in Fig.~\ref{fig:pareto_fronts_lut}.

We observe that LUT savings along the A2Q Pareto frontier come from reductions to both compute and memory resources.
The accumulator bit width affects not only the width of the register storing intermediate partial sums, but also the width of the adder circuit doing the accumulation.
Therefore, the reductions in compute resources primarily come from the reduced cost of MACs, which are directly impacted by the precision of the weights, inputs, and accumulators.
Furthermore, because FINN implements activation functions as threshold comparisons, their resource utilization exponentially grows with the precision of the accumulator and output activations~\cite{blott2018finn, umuroglu2017streamlined}.
Therefore, the reductions in memory resources are largely from the reduced storage costs of thresholds and intermediate activation buffers.

\section{Discussion}
\label{sec:discussion}

%The primary contribution of our work is A2Q, which trains QNNs for low-precision accumulation without any risk of overflow and exposes the accumulator bit width $P$ as an independent variable that can be specified by a user.
\new{The primary contribution of our work is A2Q, which trains QNNs to use low-precision accumulators during inference without any risk of overflow.
Without guaranteed overflow avoidance, one cannot reliably design hardware accelerators around low-precision accumulation.
In such scenarios, empirical estimates of overflow impact rely on \textit{a priori} knowledge of input data, which is impractical to assume in many real-world use cases.
Our guarantees allow for models and hardware to be jointly co-designed for low-precision accumulation.
A2Q exposes the accumulator bit width $P$ as an independent variable that can be specified by a user.}
Our experiments show that including the accumulator bit width in the quantization design space can improve the trade-offs between resource utilization and model accuracy.
Furthermore, while reducing the size of the accumulator invariably degrades model accuracy, using A2Q yields higher performing models than existing baselines.

It is important to highlight that our results have implications outside of the accelerators generated by FINN.
Constraining the accumulator bit width has been shown to increase inference performance on general-purpose platforms~\cite{de2020quantization,ni2021wrapnet,xie2021overflow} and reduce the compute overhead of homomorphic encryption arithmetic~\cite{lou2019she, stoian2023deep}.
Furthermore, in training QNNs for low-precision accumulation, A2Q also inherently promotes unstructured weight sparsity.
This exposes opportunities that can be exploited by both programmable hardware~\cite{colbert2021competitive, nurvitadhi2017can} as well as general-purpose processors~\cite{elsen2020fast, gale2020sparse} as the most common use of sparsity in machine learning workloads is to accelerate inference by reducing compute and memory requirements~\cite{hoefler2021sparsity}.

\textbf{Limitations.} The flexibility of FPGAs is a double-edged sword.
The bit-level control allows for the precisions of weights, activations, and now accumulators to be individually tuned for each layer in a QNN; however, this vast design space introduces a complex optimization problem.
%By exposing the accumulator bit width as an independent variable, A2Q further increases this design space.
Our study only considers uniform-precision models, but mixed-precision methods have shown promise~\cite{dong2019hawq, wang2019haq}.
State-of-the-art neural architecture search algorithms may be able to navigate this large design space more efficiently.

%\new{Finally, we observe that round-to-zero performs poorly in post-training quantization (PTQ) scenarios.
%Since A2Q relies on round-to-zero to prevent rounding errors from violating our constraints, this precludes our method from being directly applied to PTQ scenarios.
%Previous work has shown adaptive rounding can improve  PTQ accuracy over ceiling and floor rounding~\cite{nagel2020up}~\cite{nagel2020up}.
%We conjecture such approaches could be translated to .}

\new{Finally, we observe that round-to-zero performs poorly in post-training quantization (PTQ) scenarios. 
Since A2Q relies on round-to-zero to prevent rounding errors from violating our constraints, we observe poor results for A2Q in this scenario.
We conjecture that adaptive rounding techniques~\cite{nagel2020up} could alleviate the issue.}

\vspace{-0.1cm}
\section{Related Work}
\label{sec:related_work}

Another approach to training QNNs to use lower precision accumulators is to mitigate the impact of overflow on model accuracy.
%Xie \textit{et al.}~\cite{xie2021overflow} and Li~\textit{et al.}~\cite{li2022downscaling} sought to reduce the risk of overflow using an adaptive scaling factor tuned during training.
{Xie \textit{et al.}~\cite{xie2021overflow} and Li~\textit{et al.}~\cite{li2022downscaling} sought use adaptive scaling factors to reduce the expected overflow rate as empirically estimated over the training data.
Their formulation is unable to guarantee overflow avoidance as it is data dependent.}
Ni \textit{et al.}~\cite{ni2021wrapnet} proposed training QNNs to be robust to overflow using a cyclic activation function based on modulo arithmetic.
They report training instability when the overflow rate is too large, which is common when using extremely low-precision accumulators (please see Appendix~\ref{sec:motivation_exp_details} for more details).
Furthermore, both approaches model overflow using only the result of the dot product without accounting for intermediate partial sums.
Accounting for these partial sums is not easily supported by off-the-shelf deep learning frameworks nor easily generalized across target platforms.
In our work, we train QNNs to completely avoid overflow rather than simply reducing its impact on model accuracy.

Most similar to our work is that of~\cite{de2020quantization}, which proposed an iterative layer-wise optimization strategy to select mixed-precision bit widths to avoid overflow using computationally expensive heuristics.
Their derived bounds on accumulator bit width do not guarantee overflow avoidance for all edge cases and assume only signed bit widths for all data types.
Our proposed quantization method adds negligible training overhead and constrains QNNs to guarantee overflow avoidance while accounting for both signed and unsigned input data types.

Prior research has also sought to leverage weight normalization for quantization, but as a means of regularizing long-tail weight distributions during QAT~\cite{cai2019weight, li2019additive}.
Cai \textit{et al.}~\cite{cai2019weight} replace the standard $\ell_2$-norm with an $\ell_\infty$-norm and derive a projection operator to map real values into the quantized domain.
Li \textit{et al.}~\cite{li2019additive} normalize the weights to have zero mean and unit variance and observe increased stability.
In our work, we replace the $\ell_2$-norm with an $\ell_1$-norm to use the weight normalization parameterization as a means of constraining learned weights during training to use a user-defined accumulator bit width during inference.

Tangential to our work,~\cite{wang2018training} and~\cite{sakr2019accumulation} study the impact of reducing the precision of floating-point accumulators for the purpose of accelerating training.
Such methods do not directly translate to integer quantization and fixed-point arithmetic, which is the focus of this work.

\vspace{-0.1cm}
\section{Conclusion}
\label{sec:conclusion}

We present accumulator-aware quantization (A2Q), a novel quantization method designed to train QNNs for low-precision accumulation during inference.
Unlike previous work, which has sought to either reduce the risk of overflow or mitigate its impact on model accuracy, A2Q guarantees overflow avoidance and exposes the accumulator bit width as an independent variable to be specified.
To do so, A2Q constrains the $\ell_1$-norm of weights according to accumulator bounds that we derive, inherently promoting unstructured weight sparsity.
As the first principled approach to avoiding overflow, we provide theoretical motivation and derive comprehensive bounds on the accumulator bit width with finer granularity than existing literature.
We explore the use of low-precision accumulators as a means of improving the design efficiency of FPGA-based QNN inference accelerators.
Our experiments show that using our algorithm to train QNNs to use low-precision accumulators improves the trade-offs between resource utilization and model accuracy when compared to existing baselines.

\ificcvfinal
\section*{Acknowledgements}
We would like to thank Gabor Sines, Michaela Blott, Yaman Umuroglu, Nicholas Fraser, Thomas Preusser, Mehdi Saeedi, Ihab Amer, Alex Cann, Arun coimbatore Ramachandran, Chandra Kumar Ramasamy, Prakash Raghavendra, and the rest of the AMD RTG Software Technology, Edge Inference, and AECG teams for insightful discussions and infrastructure support. \\

\noindent © 2023 Advanced Micro Devices, Inc.  All rights reserved.
AMD, the AMD Arrow logo, Radeon, and combinations thereof are trademarks of Advanced Micro Devices, Inc.
Other product names used in this publication are for identification purposes only and may be trademarks of their respective companies.
\fi

\appendix

\ificcvfinal

\section*{\LARGE Appendix}

\section{Motivating Example Details}
\label{sec:motivation_exp_details}

\new{Figure~\ref{fig:accumulator_flow_diagram} illustrates a simplified abstraction of accumulation in QNN inference.
To avoid overflow, the register storing the accumulated values needs to be wide enough to not only store the result of the dot product, but also all intermediate partial sums.
Reducing the precision of the accumulator incurs a high risk of overflow which, due to wraparound two's complement arithmetic, introduces numerical errors that can degrade model accuracy~\cite{ni2021wrapnet}.}
To demonstrate the impact of overflow, we consider a 1-layer linear quantized neural network (QNN) trained to classify binary MNIST~\cite{deng2012mnist} images using 8-bit weights.
We flatten each gray-scale 28x28 image so that the inputs to the model are 784-dimensional vectors of 1-bit unsigned integers.
Keeping with our notation used throughout our paper, this translates to $N$=1, $M$=8, and $K$=784.
We train this 8-bit linear classifier using the baseline quantization-aware training (QAT) algorithm and observe a 91.5\% top-1 test accuracy when using a 32-bit accumulator.

Using our accumulator bit width data type bound, we calculate the lower bound on $P$ to be 19 bits.
Figure~\ref{fig:motivation} shows the rate of overflows per dot product grows exponentially as we reduce the accumulator bit width below this bound.
Our evaluation on the impact of overflow consists of two measures: (1) the mean absolute error on logits as measured between $P$-bit and 32-bit accumulator results; and (2) the top-1 classification accuracy of the $P$-bit accumulator result.
We observe that the increased overflow rate introduces numerical errors that proportionally increase the mean absolute error on the logits, decreasing classification accuracy.

We first evaluate the default wraparound two's complement arithmetic (\textbf{black stars}), which is known to introduce errors that degrade model accuracy~\cite{de2020quantization, li2022downscaling, ni2021wrapnet, xie2021overflow}.
Next, we evaluate na\"ively saturating values as they are accumulated (\textcolor{blue}{\textbf{blue triangles}}), which is the industry standard for avoiding overflow.
While previous work has shown that overflow ultimately degrades model accuracy due to wraparound two's complement arithmetic~\cite{de2020quantization, ni2021wrapnet}, they do not benchmark against this standard clipping solution, likely because it is expensive to model with off-the-shelf deep learning frameworks.
We show that such clipping can alleviate the accuracy degradation caused by wraparound arithmetic, but still introduce harmful errors.
Finally, we benchmark our accumulator-aware quantization (A2Q) method (\textcolor{ForestGreen}{\textbf{green circles}}) and re-train the 8-bit linear classifier from scratch using the target accumulator bit width $P$ and the same random seed.
We show that using A2Q to avoid overflow significantly improves model accuracy over both wraparound arithmetic and clipping when using extremely low-precision accumulators.
Furthermore, we find that, in our experiments, the overflow rate very quickly grows past 10\%, which is the point at which Wrapnet reports training instability~\cite{ni2021wrapnet}.
Even in those settings, A2Q maintains accuracy with respect to the floating-point counterparts, which further motivates the need to completely avoid overflow rather than simply reduce its impact on model accuracy.

\subsection{Impact of Breaking Associativity}
\label{sec:breaking_associativity}

In applying clipping, the final result of the dot product is made dependent on the order of additions, thus breaking associativity.
This can introduce non-deterministic errors when modern processors use optimizations that improve hardware utilization by re-ordering operations (\textit{e.g.}, out-of-order execution~\cite{kaeli2015heterogeneous} or cache swizzling~\cite{gale2020sparse}).
In Fig.~\ref{fig:associativity}, we show how randomly re-ordering the additions in the dot product affects the mean absolute error on the logits (left) and classification accuracy (right).
We compare modeling overflow at the outer-most loop using only the dot product result (\textcolor{red}{\textbf{red dashed line}}) against modeling overflow at the inner-most loop, which accounts for the intermediate partial sums (\textcolor{blue}{\textbf{blue histogram}}).
We also provide the baseline classification accuracy as reference (\textbf{black dashed line}).

\begin{figure}[h]
\centering
\includegraphics[width=0.93\linewidth]{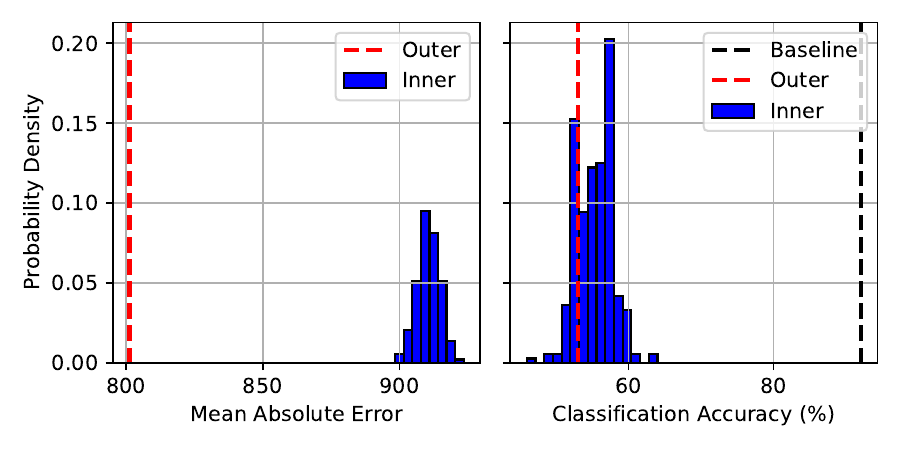}
\caption{We visualize the impact of re-ordering additions when using saturation logic on the accumulator.}
\label{fig:associativity}
\end{figure}

\section{Experiment Details and Hyperparameters}
\label{sec:exp_details}

%Here, we provide more details on the model architectures and training hyperparameters used in our experiments.
Below, we separately detail our image classification and single-image super resolution benchmarks.
For all models, we fix the input and output layers to 8-bit weights and activations for all configurations, as is common practice~\cite{gholami2021survey, hubara2017quantized, zhang2022learning}.
We also weight our regularization penalty by a constant scalar $\lambda$=1e-3 where $\mathcal{L}_\text{total} = \mathcal{L}_\text{task} + \lambda \mathcal{L}_\text{reg}$.

\begin{figure*}[t!]
\subfloat[FINN Framework Overview]{\includegraphics[width=0.6\linewidth]{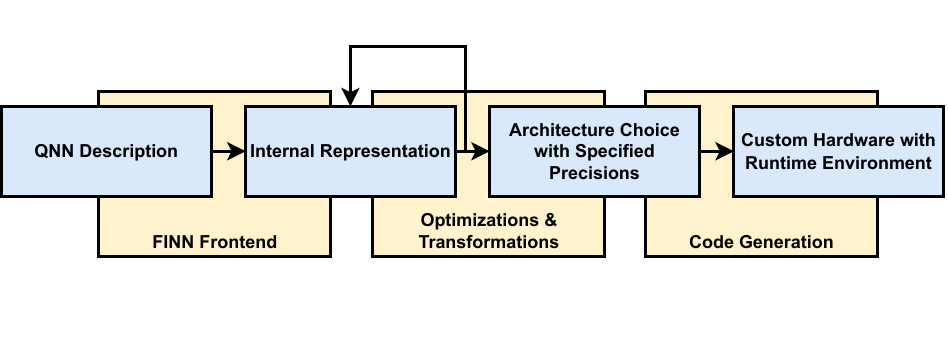}}
\subfloat[MVAU]{\includegraphics[width=0.4\linewidth]{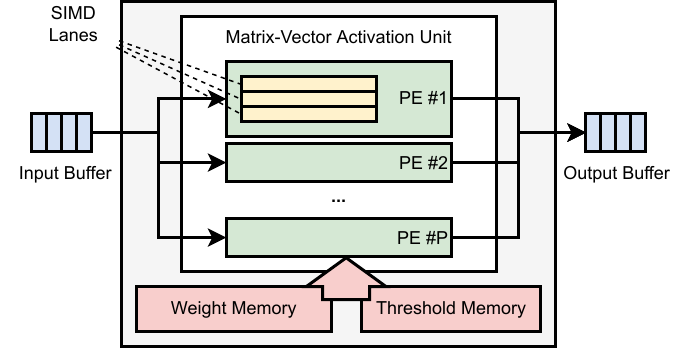}}
\caption{We adapt images from~\cite{blott2018finn, finn} to provide: (a) an overview of the FINN framework; and (b) an abstraction of the matrix-vector-activation unit (MVAU), which is one of the primary building blocks used by the FINN compiler.}
\label{fig:finn}
\end{figure*}

\subsection{Image Classification Benchmarks}
\label{sec:cifar10}

We train MobileNetV1~\cite{howard2017mobilenets} and ResNet18~\cite{he2016deep} to classify images using the CIFAR10 dataset~\cite{krizhevsky2009learning}.
We closely follow the network architectures originally proposed by the respective authors, but introduce minor variations that yield more amenable intermediate representations given the reduced image size of CIFAR10 images~\cite{krizhevsky2009learning}.
We initialize all models from floating-point counterparts pre-trained to convergence on CIFAR10 and evaluate task performance using the observed top-1 test accuracy.

\textbf{MobileNetV1.} We use a stride of 2 for both the first convolution layer and the final average pooling layer.
This reduces the degree of downscaling to be more amenable to training over smaller images.
All other layer configurations remain the same as proposed in~\cite{howard2017mobilenets}.
We use the stochastic gradient descent (SGD) optimizer to fine-tune all models for 100 epochs in batches of 64 images using a weight decay of 1e-5.
We use an initial learning rate of 1e-3 that is reduced by a factor of 0.9 every epoch.

\textbf{ResNet18.} We alter the first convolution layer to use a stride and padding of 1 with a kernel size of 3.
We remove the preceding max pool layer to reduce the amount of downscaling throughout the network.
We also use a convolution shortcut~\cite{he2016identity} rather than the standard identity as it empirically proved to yield superior results in our experiments.
All other layer configurations remain the same as proposed in~\cite{he2016deep}.
We use the SGD optimizer to fine-tune all models for 100 epochs in batches of 256 using a weight decay of 1e-5.
We use an initial learning rate of 1e-3 that is reduced by a factor of 0.1 every 30 epochs.

\subsection{Single-Image Super Resolution Benchmarks}
\label{sec:bsd300}

We train ESPCN~\cite{shi2016real} and UNet~\cite{ronneberger2015u} to upscale single images by a factor of 3x using the BSD300 dataset~\cite{MartinFTM01}.
Again, we closely follow the network architectures originally proposed by the respective authors, but introduce minor variations that yield more hardware-friendly network architectures.
We randomly initialize all models and train them from scratch.
We empirically evaluate task performance using the peak signal-to-noise ratio (PSNR) observed over the test dataset.

\textbf{ESPCN.} We replace the sub-pixel convolution with a nearest neighbor resize convolution (NNRC), which has been shown to reduce checkerboard artifacts during training~\cite{odena2016deconvolution} and can be efficiently executed during inference~\cite{ colbert2021energy}.
All other layer configurations remain the same as  proposed in~\cite{shi2016real}.
We use the Adam optimizer~\cite{kingma2014adam} to fine-tune all models for 100 epochs in batches of 16 images using a weight decay of 1e-4.
We use an initial learning rate of 1e-4 that is reduced by a factor of 0.98 every epoch.

\textbf{UNet.} We use only 3 encoders and decoders to create a smaller architecture than originally proposed by~\cite{ronneberger2015u}. 
We replace transposed convolutions with NNRCs, which are known to be functionally equivalent during inference~\cite{colbert2021energy}, but have more favorable behavior during training~\cite{odena2016deconvolution}.
We replace all concatenations with additions and reduce the input channels accordingly.
We use the Adam optimizer to fine-tune all models for 200 epochs in batches of 16 using a weight decay of 1e-4.
We use an initial learning rate of 1e-3 that is reduced by a factor of 0.3 every 50 epochs.

\section{Generating Accelerators with FINN}
\label{sec:finn}

FINN~\cite{blott2018finn,finn} is an open-source framework designed to generate custom QNN inference accelerators for AMD-Xilinx FPGAs.
For a given QNN, the FINN framework, depicted in Fig.~\ref{fig:finn}a, generates a specialized accelerator using spatial streaming dataflow architectures that are individually customized for the network topology and the data types used.
At the core of FINN is its compiler, which empowers flexible hardware-software (HW-SW) co-design by allowing a user to have per-layer control over the generated accelerator.
Weight and activation precisions can be individually specified for each layer in a QNN,
and each layer is instantiated as its own dedicated compute unit (CU). % that can be independently optimized with fine-grained parallelism.

As an example of a layer instantiated as its own CU, we provide a simplified abstraction of the matrix-vector-activation unit (MVAU) in Fig.~\ref{fig:finn}b.
The MVAU is one of the primary building blocks used for linear and convolutional layers~\cite{blott2018finn}.
Each CU consists of processing elements (PEs), which parallelize work along the output dimension, and single-instruction multiple-data (SIMD) lanes, which parallelize work along the input dimension.
%Execution over SIMDs and PEs within a layer is concurrent (\textit{i.e.}, spatial parallelism), while execution over layers within a network is pipelined (\textit{i.e.}, temporal parallelism).
All quantized monotonic activation functions in the network are implemented as threshold comparisons that map high-precision accumulated results from the preceding layer into low-precision output values.
During compilation, batch normalization, biases and even scaling factors are absorbed into this threshold logic via mathematical manipulation~\cite{umuroglu2017streamlined}.
The input and output data for the generated accelerators are streamed into and out of the chip using AXI-Stream protocols while on-chip data streams are used to interconnect these CUs to propagate intermediate activations through the layers of the network.
During inference, all network parameters are stored on-chip to avoid external memory bottlenecks.
We refer the reader to~\cite{xilinx2023finn, blott2018finn, finn} for more information.

\fi

{\small
\bibliographystyle{ieee_fullname}
\bibliography{references}
}

\end{document}